\documentclass{article} % For LaTeX2e
\usepackage[final]{colm2026_conference}
\usepackage{microtype}
\usepackage{hyperref}
\usepackage{url}
\usepackage{wrapfig}
\usepackage{booktabs}
\usepackage{graphicx}
\usepackage{caption}
\usepackage{todonotes}
\usepackage{amsmath}
\usepackage{multirow}
\usepackage{enumitem}
\setlist[itemize]{leftmargin=*}
\setlist[enumerate]{leftmargin=*}
\usepackage{algorithm}
\usepackage{arydshln} % Enables \hdashline and \cdashline
\usepackage{algpseudocode}
\usepackage{listings}
\lstdefinestyle{agenttrace}{
  basicstyle=\small\ttfamily,
  backgroundcolor=\color[gray]{0.95},
  frame=single,
  framerule=0.4pt,
  rulecolor=\color[gray]{0.7},
  xleftmargin=4pt,
  xrightmargin=4pt,
  aboveskip=6pt,
  belowskip=6pt,
  breaklines=true,
  columns=fullflexible,
  keepspaces=true,
  escapeinside={(*@}{@*)},
}

\usepackage{listings}
\usepackage{xcolor}
\usepackage{tcolorbox}

\definecolor{headerpurple}{RGB}{60,40,120}
\definecolor{lightgray}{RGB}{245,245,245}

\usepackage{minted}
\usepackage{tcolorbox}

\tcbuselibrary{minted}
\usepackage{caption}

\newtcblisting[auto counter, number within=section]{codebox}[2][]{
    listing engine=minted,
    colback=lightgray,
    colframe=headerpurple,
    title={Code~\thetcbcounter: #2},
    coltitle=white,
    colbacktitle=headerpurple,
    fonttitle=\bfseries,
    listing only,
    minted language=python,
    #1
}

\usepackage{lineno}

\definecolor{darkblue}{rgb}{0, 0, 0.5}
\hypersetup{colorlinks=true, citecolor=darkblue, linkcolor=darkblue, urlcolor=darkblue}
\newcommand{\methodabbrev}{FlexSQL}

\newcommand{\dsrsql}{DSR-SQL}
\newcommand{\reforce}{ReFoRCE}

\newcommand{\spidersnow}{Spider2-Snow}
\newcommand{\spidersqlite}{Spider2-SQLite}
\newcommand{\spidertwo}{Spider2.0}

\title{\methodabbrev: Flexible Exploration and Execution Make Better Text-to-SQL Agents}

\author{Quang Hieu Pham$^{\mathcal{A}}$ \quad
Yang He$^{\mathcal{F}}$ \quad
Ping Nie$^{\mathcal{W}}$ \quad
Canwen Xu$^{\mathcal{S}}$ \\
\textbf{Davood Rafiei}$^{\mathcal{A}}$ \quad
\textbf{Yuepeng Wang}$^{\mathcal{F}}$ \quad
\textbf{Xi Ye}\thanks{Equal advising.}$^{\,\,\,\mathcal{A}\mathcal{M}\mathcal{P}}$ \quad 
\textbf{Jocelyn Qiaochu Chen}$^{*\mathcal{A}\mathcal{M}\mathcal{N}}$ 
\\[6pt]
$^{\mathcal{A}}$University of Alberta \quad
$^{\mathcal{F}}$Simon Fraser University \quad
$^{\mathcal{W}}$University of Waterloo \quad \\
$^{\mathcal{S}}$Snowflake \quad
$^{\mathcal{P}}$Princeton University \quad 
$^{\mathcal{N}}$New York University  \\
$^{\mathcal{M}}$Alberta Machine Intelligence Institute (Amii) \quad \\
\texttt{\{quanghieupham,xi.ye,jocelyn.chen\}@ualberta.ca}
}

\begin{document}

\ifcolmsubmission
\linenumbers
\fi

\maketitle

\begin{abstract}
Text-to-SQL over large analytical databases requires navigating complex schemas, resolving ambiguous queries, and grounding decisions in actual data. Most current systems follow a fixed pipeline where schema elements are retrieved once upfront and the database is only revisited for post-hoc repair, limiting recovery from early mistakes.
We present \methodabbrev, a text-to-SQL agent whose core design principle is flexible database interaction: the agent can explore schema structure, inspect data values, and run verification queries at any point during reasoning.
\methodabbrev\ generates diverse execution plans to cover multiple query interpretations, implements each plan in either SQL or Python depending on the task, and uses a two-tiered repair mechanism that can backtrack from code-level errors to plan-level revisions. On \spidersnow, using gpt-oss-120b, \methodabbrev\ achieves a 65.4\% score, outperforming strong open-source baselines that use stronger, larger models such as gpt-o3 and DeepSeek-R1. When integrated into a general-purpose coding agent (as skills in Claude Code), our approach yields over 10\% relative improvement on Spider2-Snow. Further analysis shows that flexible exploration and flexible execution jointly contribute to the effectiveness of our approach, highlighting flexibility as a key design principle. Our code is available at: \url{https://github.com/StringNLPLAB/FlexSQL}.
\end{abstract}

\section{Introduction}

% --- Paragraph 1: Text-to-SQL over large databases is hard ---
As analytical databases grow in scale and complexity, text-to-SQL systems face an increasingly difficult reasoning problem \citep{text2sqlsurvey, text2sqlsurvey2}. The model must navigate large schemas, resolve ambiguities in the question, and ground its decisions in actual data---all before writing a single query.
Real-world data warehouses on platforms like Snowflake and BigQuery can contain hundreds of tables organized into multiple schemas \citep{bird, lei2025spider, huo2026birdinteract}, making it difficult to even identify which parts of the warehouse are relevant, let alone query them correctly. This challenge is reflected in \spidertwo \ \citep{lei2025spider}, where nearly 10\% of the 152 Snowflake databases exceed 100 tables, and the largest databases contain 60,000 to 72,000 columns.

% --- Paragraph 2: Existing pipelines and why they break ---
Most current systems follow a fixed pipeline: retrieve schema elements, generate SQL, then repair errors \citep{wang-etal-2020-rat, Li2024CodeSTB, talaei2024chess, pourreza2025chasesql, hao2025dsrsql, deng2025reforce, deepeyesql, li2025alphasql, lee-etal-2025-mcs}.
While some recent methods incorporate database interaction (e.g., executing candidate queries to check for errors, or retrieving schema elements before generation), these interactions still occur at fixed, predetermined stages.
The model explores the schema once upfront, commits to a complete query based on that static snapshot, and only revisits the database to validate or repair the output.
This rigidity means downstream repair can catch syntax errors, but it cannot recover when the model chose the wrong tables or misunderstood a column's meaning in the first place.

% --- Paragraph 3: Introducing AutoPylot — flexible interaction as the core principle ---
We present \methodabbrev, a text-to-SQL agent whose core design principle is \emph{flexible database interaction}: the agent incrementally discovers schema structure, grounds its decisions in actual data values, and can revise its approach based on what it finds---at any point during reasoning.
For instance, if the agent encounters an error during code generation that stems from a wrong table choice, it can backtrack all the way to schema exploration rather than being trapped in a repair loop.

% --- Paragraph 3b: Tools that enable flexible interaction ---
To make this flexible interaction concrete, \methodabbrev\ equips the agent with a suite of tools designed specifically for incremental database exploration: the agent can navigate schema hierarchies top-down, inspect actual column values, and run exploratory queries against the database.
These tools are available throughout the entire reasoning process, so the agent can ground its plans in real data and verify intermediate results at any point.

% --- Paragraph 4: Ambiguity requires diversity; varied strategies require flexibility ---
While these tools address the grounding problem, a second challenge remains: natural language queries are often inherently ambiguous, admitting multiple valid interpretations \citep{10.14778/2735461.2735468, pourreza-rafiei-2023-evaluating, klopfenstein2026spotit}.
Rather than committing to one, \methodabbrev\ uses diversity-enforced sampling \citep{zhang2025verbalized} to produce $K$ plans covering different table choices, join paths, and predicate readings.
By exploring a broader hypothesis space, \methodabbrev\ naturally improves its robustness against query ambiguity, and effectively leverages inference-time compute as pass@$K$ improves with more candidates \citep{wang2023selfconsistency, XiYanSQL}.

The diverse plans also call for flexibility in implementation: some strategies involve multi-step transformations that are cumbersome in pure SQL, so \methodabbrev\ allows each plan to be implemented in either SQL or Python. Python's more flexible control structures are especially useful for problems that require iterative refinement and branching conditionals \citep{transpilation-with-llm}, and Python implementations are subsequently translated into SQL.
Together, flexible interaction, diverse planning, and bilingual generation allow \methodabbrev\ to cover a broad space of candidate solutions while grounding each one in the actual database.

% --- Paragraph 5: Results preview ---
We evaluate \methodabbrev\  using the open-source \texttt{gpt-oss} models \citep{gptoss} on two settings from the \spidertwo \ benchmark \citep{lei2025spider}, covering both cloud data warehouses (\spidersnow) and  local databases (\spidersqlite). With \texttt{gpt-oss-120b}, \methodabbrev\ reaches 65.44\% Majority@8 on Spider-Snow, outperforming \dsrsql\ with DeepSeek-R1 and \reforce\ with gpt-o3 despite using a significantly smaller backbone. Under our framework, \texttt{gpt-oss-20b} also either surpasses or remains comparable to baselines using \texttt{gpt-oss-120b}. We further integrate \methodabbrev\ as skills in a general-purpose coding agent (e.g., Claude Code \citep{claude_code}), yielding over 10\% relative improvement on \spidersnow. 
% \inlinetodo{Mention we use two splits both snow and sqlite}
% \inlinetodo{Fill in final numbers and model names.}

% --- Paragraph 6: Contributions ---
Our contributions are:
\begin{itemize}
    \item A text-to-SQL agent built around flexible database interaction, equipped with a suite of tools for incremental schema navigation, data inspection, and code execution that the agent can use throughout the entire reasoning process.
    \item A two-stage plan-to-program architecture with a flexible backtracking mechanism that allows the agent to revise not only its code but also its plan and schema assumptions when errors are detected.
    \item Diversity-enforced plan sampling to handle query ambiguity by covering multiple valid interpretations, combined with bilingual program generation (SQL or Python) to flexibly implement the varied strategies that arise.
    \item Strong empirical results on \spidertwo, where \methodabbrev\ consistently outperforms baselines across models and exhibits effective test-time scaling with increased sampling budgets.
\end{itemize}

\section{Motivating Example}
\label{sec:motivating-example}

\begin{figure*}[t]
\centering
\includegraphics[width=\textwidth,trim={25pt 300pt 320pt 130pt},clip]{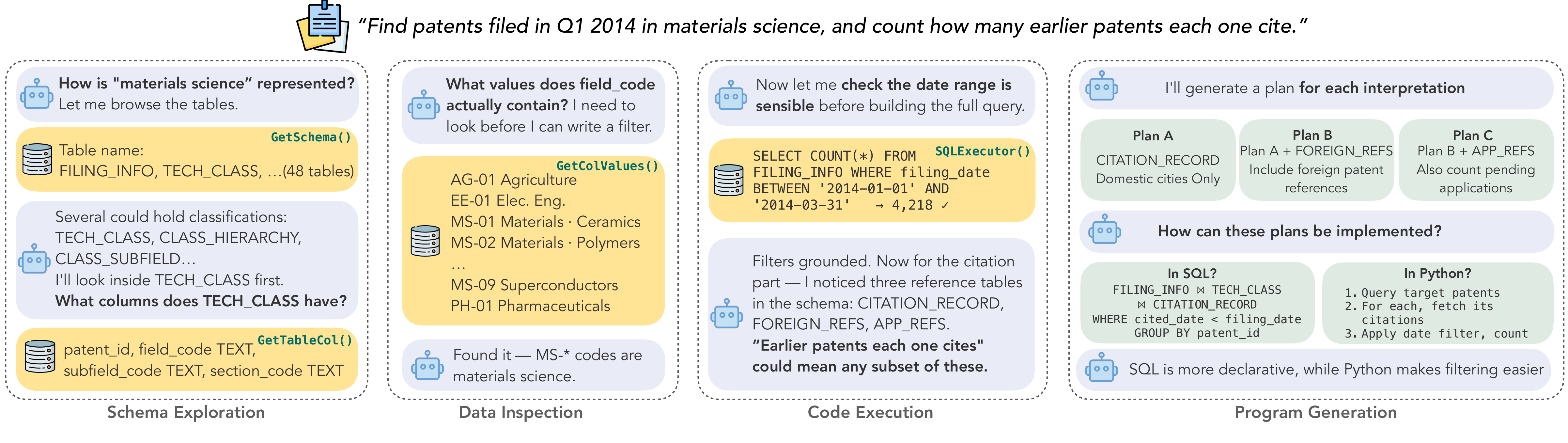}
\caption{Condensed trace of \methodabbrev\ on the motivating example. The agent progressively discovers the schema, grounds its filters in actual data values, verifies intermediate results, and generates diverse plans covering different interpretations of the query---all through incremental tool interaction with the database.}
\label{fig:agent-trace}
\end{figure*}

% --- Setup: the query and database ---
We illustrate the key challenges of text-to-SQL over complex databases and how \methodabbrev\ addresses them through a concrete example.
Consider the following question posed over an intellectual property database containing 48 tables spanning filings, inventors, technology classifications, citations, and legal events:
% \smallskip
\emph{``Find patents filed in Q1 2014 in materials science, and count how many earlier patents each one cites.''}
% \smallskip

\noindent Answering this question requires solving two problems: identifying how ``materials science'' is represented in the database, and deciding what ``earlier patents each one cites'' means.
Both turn out to be harder than they appear.

% --- Challenge 1: Schema discovery and data grounding ---
\paragraph{Challenge 1: Schema discovery and data grounding.}
The phrase ``materials science'' does not appear as a table or column name in the database.
The relevant information is buried in the \texttt{field\_code} column of a table called \texttt{TECH\_CLASS}, one of 48 tables, several of which have plausible-sounding names like \texttt{CLASS\_HIERARCHY}, \texttt{CLASS\_SUBFIELD}, and \texttt{FIELD\_SECTION}.
Even after locating the right table, the correct filter (\texttt{field\_code IN ('MS-01', ..., 'MS-09')}) can only be determined by examining actual data values, not schema metadata.
A system that performs schema linking once upfront and never revisits the database may fail to discover this mapping.
In practice, we observe that one baseline missed \texttt{TECH\_CLASS} entirely, returning all patents filed in Q1~2014 instead of only those in materials science.

% Because all sampled candidates shared the same schema selection, majority voting reinforced rather than corrected the error.

% --- Challenge 2: Query ambiguity ---
\paragraph{Challenge 2: Query ambiguity.}
The phrase ``earlier patents each one cites'' admits multiple valid readings.
The database contains three citation-related tables (\texttt{CITATION\_RECORD}, \texttt{FOREIGN\_REFS}, \texttt{APP\_REFS}), and it is unclear which subset the query intends: (1) using only domestic citations, (2) including foreign references, or (3) also counting pending applications, each yielding different results.
A second baseline found \texttt{TECH\_CLASS} correctly but unioned all three citation tables, overcounting by an order of magnitude.
Again, all candidates shared the same structural choice, leaving no room for voting to recover.

% --- How AutoPylot handles the same query ---
\paragraph{How \methodabbrev\ handles this query.}
Figure~\ref{fig:agent-trace} shows a condensed trace of the agent's interaction with the database.
Rather than committing to a fixed schema selection, the agent explores incrementally: it calls \texttt{GetSchema} to list all 48 tables, identifies \texttt{TECH\_CLASS} as a candidate, inspects its columns via \texttt{GetTableCol}, and then calls \texttt{GetColValues} to discover that \texttt{field\_code} values \texttt{MS-01} through \texttt{MS-09} correspond to materials science.
With the technology domain filter grounded, the agent also needs to verify the date constraint, which is stored in a separate table, \texttt{FILING\_INFO}.
Before writing the full query, it runs a verification query (\texttt{SELECT COUNT(*) FROM FILING\_INFO} with the date filter) to confirm the date range yields a reasonable number of rows.

% --- Diverse plans ---
With the schema grounded, the agent turns to the ambiguity in ``earlier patents.''
Rather than committing to one interpretation, it generates multiple plans that differ in which citation tables to use and how to apply date filters: Plan~A uses only \texttt{CITATION\_RECORD} with a filter on \texttt{cited\_date}; Plan~B additionally includes \texttt{FOREIGN\_REFS}; Plan~C further adds \texttt{APP\_REFS}.
Each plan can also be implemented in either SQL or Python. For instance, a plan that iterates over each patent's citations and applies per-patent filtering is more naturally expressed as a Python loop than as a multi-way self-join in SQL.
Because all plans discover \texttt{TECH\_CLASS} through tools, they agree on which patents to retrieve but produce different citation counts; majority voting over execution outputs selects the most supported interpretation.

The rest of this paper describes the three mechanisms at work: tool-based database interaction (\S\ref{sec:tools}), diversity-enforced plan generation (\S\ref{sec:plan-gen}), and program generation(\S\ref{sec:prog-gen}).

\section{Tool Interfaces for Database Querying}
\label{sec:tools}

We design a set of tools that allow language model agents to interact with relational databases incrementally during text-to-SQL reasoning.
Agents in \methodabbrev\ are initialized with only coarse-grained metadata (e.g., a list of schema names) and use these tools to retrieve information on demand as their understanding of the query develops.
The tools are organized around three capabilities: navigating schema structure, inspecting data values, and executing code against the database.

\paragraph{Database hierarchy and agent inputs.}
Analytical databases on platforms like Snowflake and BigQuery are typically organized hierarchically: a \emph{database} contains one or more \emph{schemas}, each grouping a set of related \emph{tables} by business domain.
For example, a city government database might have a \texttt{Transportation} schema (with tables for bike-sharing and traffic) and an \texttt{Environment} schema (with tables for air quality and waste management).
Our tools are designed around this hierarchy.\footnote{For databases without an explicit schema layer (e.g., standard relational databases), the schema level is simply flattened and the agent navigates directly from the database to its tables.}
At initialization, the agent receives the user query, any available external knowledge documents, and a list of schema names in the database.
All further information, such as table names, column definitions, and data values, is retrieved through tool calls.

\paragraph{Schema exploration.}
The first two tools let the agent navigate this hierarchy top-down:
\begin{itemize}
    \item \textbf{\texttt{GetSchema(schema\_name)}}: Given a schema name (e.g., \texttt{Transportation}), returns all table names within that schema. This is useful because a single database may contain dozens of schemas, and the agent needs to narrow its focus to the relevant domain before examining individual tables.
    \item \textbf{\texttt{GetTableCol(table\_name)}}: Given a fully qualified table name, returns its column names, data types, and a few sample values. This provides enough information to assess whether a table is relevant and how it might join with other tables.
\end{itemize}

\paragraph{Data inspection.}
Schema metadata alone is often insufficient. As illustrated in Section~\ref{sec:motivating-example}, the agent may examine actual data values to resolve ambiguities. Two tools support this:
\begin{itemize}
    \item \textbf{\texttt{GetColValues(column\_name, table\_name)}}: Returns the distinct values stored in a specific column. This is essential when the query depends on categorical values that are not apparent from column names alone (e.g., discovering that ``chemistry'' corresponds to classification codes \texttt{C05}--\texttt{C13}).
     \item \textbf{\texttt{FindRows(term, column\_name, table\_name, additional\_columns)}}: Performs a keyword search within a column and returns matching rows, optionally including values from other columns in the same table. Here, \texttt{term} is the search string (e.g., a product name or abbreviation) the agent wants to locate in the data. This helps verify whether a value exists and how it relates to other columns before committing to a filter condition.
\end{itemize}

\paragraph{Code execution.}
The final two tools let the agent run code against the database, supporting iterative development and verification of partial solutions:
\begin{itemize}
    \item \textbf{\texttt{SQLExecutor(sql\_query)}}: Executes a SQL query and returns the result set or an error message. The agent can use this to test subqueries, verify join logic, or check intermediate results before assembling a full query.
    \item \textbf{\texttt{PythonExecutor(program)}}: Executes Python code in a stateful sandbox with database access. This supports multi-step reasoning, such as fetching data with a simple query, transforming it programmatically, and verifying the output, which can be difficult to express in a single SQL statement.
\end{itemize}

\begin{figure}[t]
\begin{center}
\includegraphics[width=0.9\linewidth,trim={140pt 340pt 200pt 285pt},clip]{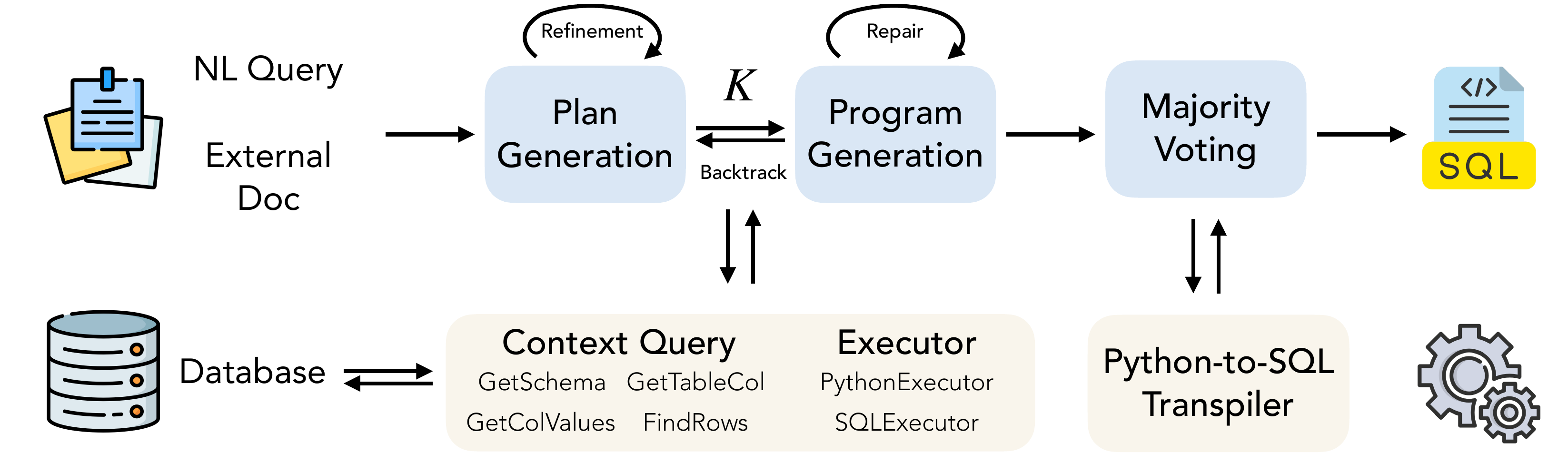}
\caption{Overview of the \methodabbrev\ framework. The agent takes a natural language query and optional external documents as input, and produces a SQL query through three components: \textbf{Plan Generation} explores the database via context query and executor tools to produce $K$ diverse plans, with a refinement loop; \textbf{Program Generation} translates each plan into SQL or Python, with a repair loop for code-level errors and backtracking to Plan Generation for plan-level errors; \textbf{Majority Voting} selects the consensus answer and transpiles Python solutions to SQL when needed.}
\label{fig:overview}
\end{center}
\end{figure}

\section{\methodabbrev\ Framework}
\label{sec:framework}

Given a user query $q$, a database $\mathcal{D}$ with schemas $\{S_1, \dots, S_n\}$, and optional domain documents $\mathcal{K}$, \methodabbrev\ produces a final SQL query $a^*$ and its execution result $o^*$.
As shown in Figure~\ref{fig:overview}, the framework has three components.
\textbf{Plan Generation} (\S\ref{sec:plan-gen}) uses the context query tools (\texttt{GetSchema}, \texttt{GetTableCol}, \texttt{GetColValues}, \texttt{FindRows}) and executors (\texttt{SQLExecutor}, \texttt{PythonExecutor}) to explore the database and construct $K$ diverse execution plans, with a refinement loop that iterates on each plan using tool feedback.
\textbf{Program Generation} (\S\ref{sec:prog-gen}) translates each plan into executable SQL or Python, with a repair loop for code-level errors and a backtracking mechanism that revises the plan itself when deeper reasoning errors are detected.
Finally, \textbf{majority voting} groups all programs by execution output regardless of language, selects the consensus answer, and transpiles any Python solution to SQL.

\subsection{Plan Generation}
\label{sec:plan-gen}

The first stage produces $K$ execution plans that cover different interpretations of the query.
Each plan is a natural language description of a query strategy, e.g., which tables to use, how to join them, what filters to apply, grounded in the actual database through tool interaction.

\paragraph{Context building.}
Large analytical databases can contain hundreds of tables, many irrelevant to any given query.
Before planning, we apply lightweight preprocessing: pruning columns that are entirely null and grouping tables that share identical schemas but differ only by a time-based suffix (e.g., \texttt{GA\_SESSION\_20240101}, \texttt{GA\_SESSION\_20240202}).
If external documents $\mathcal{K}$ are provided, we extract a query-relevant summary.
Full preprocessing details are provided in Appendix~\ref{app:context_building}.

Even after preprocessing, the database may remain too large for direct context grounding.
We address this with a routing step: the agent examines the list of schema names and uses \texttt{GetSchema} and \texttt{GetTableCol} to identify a relevant subset of schemas $\mathcal{S}_{\text{rel}} \subseteq \mathcal{D}$, exploiting the database $\to$ schema $\to$ table hierarchy described in Section~\ref{sec:tools}.

\paragraph{Diversity-enforced sampling.}
Given $\mathcal{S}_{\text{rel}}$, the agent generates $K$ plans in parallel batches of size $M$ using verbalized sampling \citep{zhang2025verbalized}: each batch is prompted to produce plans that differ from those already generated, encouraging coverage of genuinely different query interpretations.
During planning, the agent actively explores the database using the full tool suite such as inspecting schema structure, examining column values, and running test queries, to ground its plans in the actual data.

\paragraph{Plan refinement.}
Each generated plan is reviewed by the agent against the information gathered through tool interaction.
If the review identifies issues (e.g., a referenced table does not exist, or a filter contradicts the actual column values), the agent refines the plan by re-exploring the database and regenerating, as shown by the refinement loop in Figure~\ref{fig:overview}.

\subsection{Program Generation}
\label{sec:prog-gen}

The second stage translates each plan into an executable program, revises it if needed, and selects the final answer.

\paragraph{Bilingual program synthesis.}
For each plan $p_i$, the agent generates a program $a_i$ in either Python or SQL, choosing the language that best fits the task.
SQL is well-suited for declarative relational queries, while Python is better suited for procedural workflows involving multi-step transformations, iterations, or computations that are cumbersome in a single SQL statement.
During synthesis, the agent can use \texttt{SQLExecutor} and \texttt{PythonExecutor} to test partial implementations by running subqueries, checking intermediate outputs, and debugging, before committing to a final program.
\paragraph{Repair.}
After execution, a review step evaluates whether the program's output $o_i$ is consistent with the original query and plan.
If a \emph{code-level} error is detected, such as syntactic errors, wrong aggregation, or incorrect column references, the program is regenerated with the error message as feedback (see the repair loop in Figure~\ref{fig:overview}).
We allow up to $R$ repair rounds (set to 3 in our experiments).

\paragraph{Plan backtracking.}
If the review instead identifies a \emph{plan-level} error, such as wrong tables, missing joins, or an incorrect interpretation of the query, repair alone cannot fix the problem.
In this case, the system backtracks to Plan Generation (Figure~\ref{fig:overview}): the agent revises the plan itself, potentially re-exploring the database with tools, before re-attempting program synthesis.
This allows the system to recover from deeper reasoning errors, not just surface-level implementation mistakes.

\paragraph{Majority voting and transpilation.}
\label{sec:voting}
After all $K$ programs have been generated---some in SQL, some in Python---we group them by execution output regardless of language and select the largest equivalence class $\mathcal{A}^*$ as the consensus answer.
This allows SQL and Python solutions to reinforce each other during voting, which is important because some query strategies are easier to implement correctly in Python.
If the consensus answer was produced by a SQL query, we return it directly; otherwise, we transpile a Python program from $\mathcal{A}^*$ into SQL so that the final output is always a SQL query.
The transpiler generates a candidate query, executes it, and verifies that its output matches the consensus before returning. We include our heuristic transpilation rules in Appendix~\ref{app:heuristic-transpilation-rules}.

\section{Experiments}

\paragraph{Datasets.} We evaluate \methodabbrev\ on two subsets of \spidertwo\ \citep{lei2025spider}: \spidersnow\ (544 questions over Snowflake databases) and \spidersqlite\ (135 questions over SQLite databases). While \spidersnow\ contains enterprise-scale databases spanning hundreds of tables, \spidersqlite\ features moderately sized schemas averaging 14.4 tables (with a maximum of 38). Together, this benchmark suite tests performance across distinct SQL dialects and varying schema complexities.

\paragraph{Baselines.} We evaluate our method against two strong open-source baselines on the \spidertwo\ leaderboard: \reforce\ \citep{deng2025reforce} and \dsrsql\ \citep{deng2025reforce}. For a fair comparison, all experiments use \texttt{gpt-oss-120b} and \texttt{gpt-oss-20b} \citep{gptoss} with identical sampling and repair budgets and equal access to database tools. At the time we develop 
\methodabbrev, only the two baselines provided open-source codebases. Our inference configurations are in Appendix~\ref{sec:gen_config}. We also demonstrate the broader generalization of \methodabbrev \ by integrating our flexible interaction paradigm to Claude Code, a widely adopted general-purpose coding agent.

\paragraph{Evaluation metrics.} We consider three evaluation dimensions. (1) \textbf{Execution accuracy}, measured by Pass@1 against the released golden execution results. (2) \textbf{Test-time scalability}, assessed via Pass@8 and majority-vote accuracy over eight samples (\textbf{Majority@8}). (3) \textbf{Answer diversity}, measured by \textbf{micro accuracy}: since \spidertwo\ includes multiple gold answers per question, micro accuracy assigns fractional credit (e.g., $1/N$ for $N$ gold answers) based on whether the model recovers any valid solution. We also evaluate schema linking quality at the table level using precision and recall.

\subsection{Main results}
\label{sec:main_results}

\begin{table*}[t]
\scriptsize
\centering
\resizebox{\textwidth}{!}{%
\begin{tabular}{ll ccc ccc}
\toprule
\multirow{2}{*}{\textbf{Model}} & \multirow{2}{*}{\textbf{Metric}} & \multicolumn{3}{c}{\textbf{\spidersnow}} & \multicolumn{3}{c}{\textbf{\spidersqlite}} \\
\cmidrule(lr){3-5} \cmidrule(lr){6-8}
& & \textbf{\dsrsql} & \textbf{\reforce} & \textbf{Ours} & \textbf{\dsrsql} & \textbf{\reforce} & \textbf{Ours} \\
\midrule
\multirow{3}{*}{gpt-oss-120b} 
& Pass@1 & 33.27 & 44.12 & \textbf{55.15} & 48.15 & 45.19 & \textbf{57.78} \\
& Majority@8 & 50.37 & 48.90 & \textbf{59.74} & 51.85 & 54.07 & \textbf{64.44} \\
& Pass@8 & 63.24 & 62.32 & \textbf{78.68} & 57.78 & 71.11 & \textbf{78.52} \\
& Micro Acc. & 28.38 & 28.11 & \textbf{44.07} & 25.79 & 32.11 & \textbf{49.69} \\
\midrule
\multirow{3}{*}{gpt-oss-20b} 
& Pass@1 & 32.54 & 36.76 & \textbf{43.20} & 34.81 & 42.96 & \textbf{50.37} \\
& Majority@8 & 42.65 & 43.01 & \textbf{50.92} & 37.04 & 46.67 & \textbf{54.07} \\
& Pass@8 & 52.39 & 59.01 & \textbf{71.14} & 39.26 & 68.15 & \textbf{78.52} \\
& Micro Acc. & 18.55 & 24.91 & \textbf{37.85} & 20.44 & 29.52 & \textbf{45.34} \\
\bottomrule
\end{tabular}%
}
\caption{Performance comparison (in \%) on \spidersnow \ and \spidersqlite \ datasets.}
\label{tab:main_results_compact}
\end{table*}

% ----------------- leaderboard + scaling side by side ---------------------- %
\begin{figure*}[t]
\centering
\begin{minipage}[c]{0.35\textwidth}
\centering
\includegraphics[width=\linewidth]{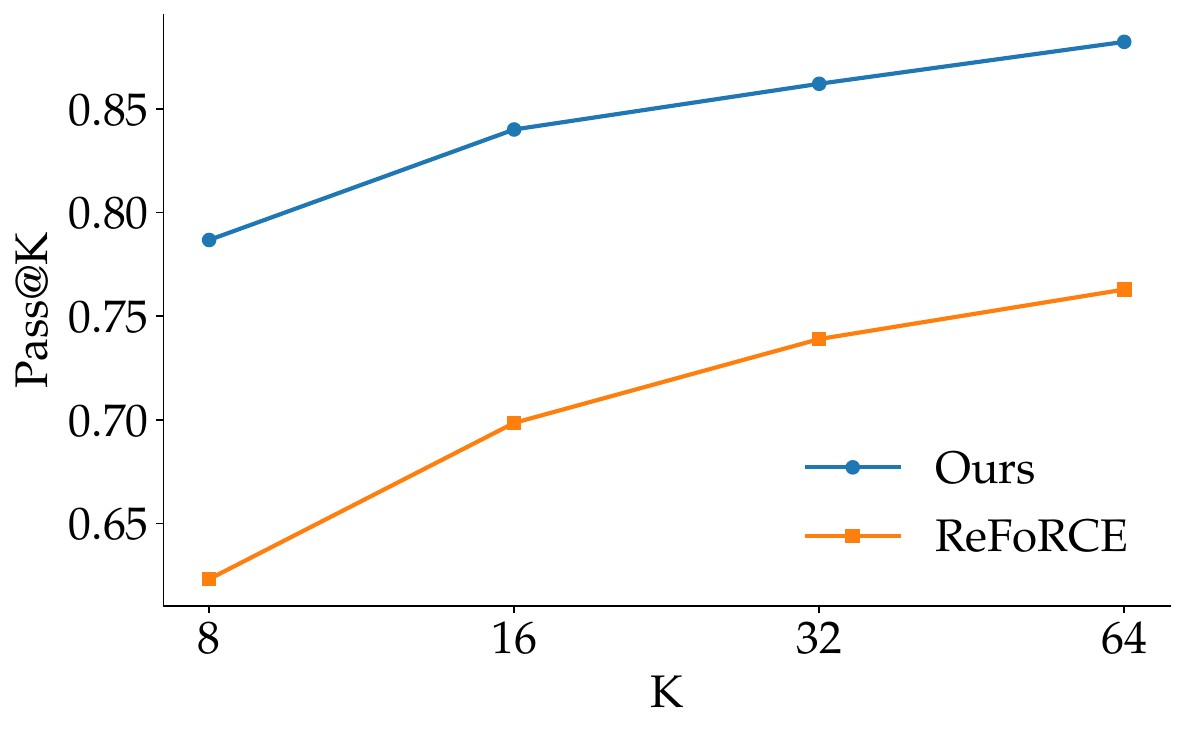}
\captionof{figure}{Pass@$K$ test-time scaling of \methodabbrev\ and \reforce.}
\label{fig:scaling_performance}
\end{minipage}%
\hfill
\begin{minipage}[c]{0.55\textwidth}
\centering
\resizebox{\linewidth}{!}{%
\small
\begin{tabular}{ll c}
\toprule
\textbf{Method} & \textbf{Model} & \textbf{Majority@K} \\
\midrule
\dsrsql
& DeepSeek-R1 & 63.80 \\
\midrule
\reforce
& gpt-o3 & 62.89 \\
\midrule
\multirow{3}{*}{\textbf{Ours}}
& gpt-oss-20b, $K=8$ & 50.92 \\
& gpt-oss-120b, $K=8$ & 59.74 \\
& gpt-oss-120b, $K=16$ & \textbf{65.44} \\
\bottomrule
\end{tabular}%
}
\captionof{table}{Majority@K on \spidersnow\ compared against leading open-source systems on the \spidertwo\ leaderboard.}
\label{tab:snow_leaderboard}
\end{minipage}
\end{figure*}

\paragraph{Overall performance.}
Table~\ref{tab:main_results_compact} compares \methodabbrev\ against \dsrsql\ and \reforce. Across both \spidersnow\ and \spidersqlite, our method consistently achieves higher execution accuracy (Pass@1), test-time scalability (Pass@8, Majority@8), and answer diversity (micro accuracy) within each model size class. With \texttt{gpt-oss-120b}, \methodabbrev\ reaches a Pass@1 of 55.15\% on \spidersnow\ and 57.78\% on \spidersqlite, yielding absolute improvements of 11.03\% and 9.63\% over the strongest respective baselines.

\paragraph{Parameter efficiency.}
\methodabbrev\ remains competitive even with the smaller \texttt{gpt-oss-20b} backbone. On \spidersqlite, our 20b model achieves a Pass@1 of 50.37\%, outperforming both 120b baselines (\reforce: 45.19\%, \dsrsql: 48.15\%). On \spidersnow, our 20b Pass@1 of 43.20\% remains comparable to the strongest 120b baseline (44.12\%).

\paragraph{Test-time scaling.}
Figure~\ref{fig:scaling_performance} shows Pass@$K$ as we scale the sampling budget from $K{=}8$ to $K{=}64$ with \texttt{gpt-oss-120b} on \spidersnow. \methodabbrev\ maintains a consistent lead over \reforce\ across the entire range, confirming that our approach delivers reliable gains as test-time compute scales.

\paragraph{Leaderboard comparison.}
To quantify the practical impact of this scaling behavior, Table~\ref{tab:snow_leaderboard} compares \methodabbrev\ against leading open-source entries on the \spidersnow\ leaderboard, including \dsrsql\ with DeepSeek-R1 and \reforce\ with gpt-o3. Our base \texttt{gpt-oss-120b} model with $K{=}8$ already achieves a competitive 59.74\%. Scaling to $K{=}16$ yields 65.44\%, surpassing both \dsrsql\ with DeepSeek-R1 (63.80\%) and \reforce\ with gpt-o3 (62.89\%). This demonstrates that \methodabbrev\ can match or exceed open-source systems that use stronger reasoning models.

\subsection{Analysis}

To understand the contribution of each design choice, we ablate bilingual generation, diversity-enforced planning, and flexible repair on the \spidersqlite\ split using both \texttt{gpt-oss-120b} and \texttt{gpt-oss-20b} (Table~\ref{tab:ablation_sqlite_detailed}).

\paragraph{Effectiveness of diversity-enforced planning.}
We ablate the verbalized diversity constraint by letting the agent generate $K$ independent plans without enforcing variation. This reduces Majority@8 from 64.44\% to 55.56\% for the 120b model and from 54.07\% to 49.63\% for the 20b model, confirming that explicitly prompting for varied interpretations improves majority-vote quality. By generating plans in batches and enforcing diversity, the agent avoids converging on a single, potentially flawed interpretation, giving majority voting a broader set of candidates to draw consensus from. Interestingly, for the 120b model, removing diverse planning increases Pass@8 from 78.52\% to 83.70\%, revealing an exploration--exploitation trade-off: enforcing diversity strengthens majority consensus but can occasionally lead the model to overcomplicate its reasoning at the expense of simpler, viable answers.

\begin{table*}[t]
\centering
\begin{minipage}[c]{0.52\textwidth}
\centering
\resizebox{\linewidth}{!}{%
\begin{tabular}{lccc}
\toprule
\textbf{Model Setup} & \textbf{Majority@8} & \textbf{Pass@8} & \textbf{Micro Acc.} \\
\midrule
\textbf{gpt-oss-120b} & \textbf{64.44} & 78.52 & 49.69 \\
\quad w/o Python (SQL Only) & 52.59 & 68.89 & 41.10 \\
\quad w/o diverse planning & 55.56 & \textbf{83.70} & \textbf{50.23} \\
\quad w/o plan backtracking & 61.48  &  80.00   & 48.64 \\
\quad w/o all repair & 57.04 & 79.26 & 46.79 \\
\midrule
\textbf{gpt-oss-20b} & \textbf{54.07} & \textbf{78.52} & \textbf{45.34} \\
\quad w/o Python (SQL Only) & 42.22 & 60.00 & 36.08 \\
\quad w/o diverse planning & 49.63 & 74.07 & 43.32 \\
\quad w/o plan backtracking & 52.59  & 73.33 & 42.21  \\
\quad w/o all repair & 51.85 & 75.37 & 44.39 \\
\bottomrule
\end{tabular}%
}
\caption{Ablation study on \spidersqlite\ (in \%) evaluating the impact of the Python interpreter, diverse planning, and repair.}
\label{tab:ablation_sqlite_detailed}
\end{minipage}%
\hfill
\begin{minipage}[c]{0.45\textwidth}
\centering
\resizebox{\linewidth}{!}{%
\begin{tabular}{lccc}
\toprule
\textbf{Method} & \textbf{Precision} & \textbf{Recall} & \textbf{F1} \\
\midrule
\textbf{Ours} + oss-120b (Best of 8) & \textbf{95.46} & 95.06 & \textbf{95.26} \\
\textbf{Ours} + oss-120b (First plan) & 87.26 & 89.13 & 88.19 \\
CHESS + oss-120b & 69.21 & 73.36 & 71.22\\
\reforce\ + o4 + o3 + o4 mini \textsuperscript{*} & 66.83 & \textbf{99.73} & 80.03 \\
\reforce\ + DeepSeek-V3\textsuperscript{\dag} & 58.09 & 62.59 & 60.26 \\
\dsrsql\ + DeepSeek-V3\textsuperscript{\dag} & 75.62 & 91.13 & 82.65 \\
\bottomrule
\end{tabular}%
}
\caption{Table-level schema linking on \spidersnow. \emph{Best of 8}: best schema across $K{=}8$ plans. \emph{First plan}: schema from the first plan only. \textsuperscript{*}From \citet{deng2025reforce} GitHub release. \textsuperscript{\dag}From \citet{hao2025dsrsql}.}
\label{tab:table_level_sl_perf}
\end{minipage}
\end{table*}

\paragraph{The role of bilingual generation.}
\label{sec:analysis}
% \begin{table}[h]
\begin{wraptable}{r}{0.55\textwidth}
\vspace{-1em}
\centering
\resizebox{\linewidth}{!}{
\begin{tabular}{llccc}
\toprule
\textbf{Dataset} & \textbf{Model} & \textbf{Only Python} & \textbf{Only SQL} & \textbf{Both} \\
\midrule
\multirow{2}{*}{\spidersnow} & \textbf{120b} & 7.1\% & 48.8\% & 44.1\% \\
 & \textbf{20b} & 5.5\% & 61.0\% & 33.5\% \\
\midrule
\multirow{2}{*}{\spidersqlite} & \textbf{120b} & 27.6\% & 25.3\% & 47.1\% \\
 & \textbf{20b} & 23.3\% & 6.8\% & 69.9\% \\
\bottomrule
\end{tabular}
}
\caption{Breakdown of solved questions by language for Majority@8 groups on \spidersnow\ and \spidersqlite. SQL/Python ratio within ``Both'' is in Appendix~\ref{app:gen_ratio}.}
\label{tab:py_sql_solvability}
\vspace{-1em}
\end{wraptable}

As shown in Table~\ref{tab:ablation_sqlite_detailed}, removing the Python interpreter (``SQL Only'') causes the largest drop across all ablations: Majority@8 falls from 64.44\% to 52.59\% for the 120b model and from 54.07\% to 42.22\% for the 20b model.

Table~\ref{tab:py_sql_solvability} provides a finer-grained view by categorizing solved questions according to which languages produced correct answers. A large proportion fall into the ``Both'' category (47.1\% for the 120b model and 69.9\% for the 20b model on \spidersqlite), indicating that Python consistently generates viable solutions alongside SQL. Moreover, Python exclusively solves up to 27.6\% of queries on \spidersqlite. Together, these results show that the two languages are complementary: the bilingual toolkit expands coverage by providing multiple pathways to the correct answer.

\paragraph{The impact of repair and backtracking.}
Disabling the repair mechanism, which removes both code-level repair and plan-level backtracking, reduces Majority@8 from 64.44\% to 57.04\% (120b) and from 54.07\% to 51.85\% (20b).
Without repair, the system can neither fix implementation errors (e.g., wrong aggregation or column references) nor backtrack to the planning stage when deeper reasoning errors are detected (e.g., wrong table selection or missing joins).

\paragraph{Schema linking performance.}
% Table 5 (schema linking) is now placed side-by-side with Table 3 above.

Table~\ref{tab:table_level_sl_perf} reports table-level schema linking on \spidersnow. Existing approaches retrieve schema upfront: \reforce\ and \dsrsql\ include all candidate tables and fall back to independent table filtering on overflow, while CHESS \citep{talaei2024chess} applies column-level filtering followed by table identification, incurring the high computational cost of a separate language model call for each column. In contrast, \methodabbrev\ explores the database incrementally and retrieves schema elements on demand, yielding substantially higher precision with comparable recall and the best overall F1. Even the first generated plan attains competitive F1, suggesting that flexible exploration can reduce context overhead while preserving accurate schema grounding.

\begin{wraptable}{r}{0.55\textwidth} % 'r' for right, '0pt' for auto-width
% \vspace{em}
\vspace{-1em}
\centering
\resizebox{\linewidth}{!}{%
\begin{tabular}{lcc}
\toprule
\textbf{Source Model} & \textbf{\spidersnow} & \textbf{\spidersqlite} \\ 
\midrule
gpt-oss-20b & 15/15 & 17/17 \\ 
gpt-oss-120b & 22/23 & 23/24 \\ 
gpt-oss-120b (pass@16 (MV)) & 13/14 & -- \\ 
\midrule
% \midrule
Transpiled with gpt-oss-20b & \multicolumn{2}{c}{81/93 $\approx 87.09\%$}  \\
\textbf{Overall accuracy} & \multicolumn{2}{c}{90/93 $\approx 96.77\%$} \\
\bottomrule
\end{tabular}
}
\caption{Transpilation accuracy of our framework on all experiments with \texttt{gpt-oss} models on \spidersnow \ and \spidersqlite.}
\label{tab:python-sql-transpilation}
\vspace{-1em}
\end{wraptable}

\paragraph{Python to SQL accuracy.}
\label{sec:py_to_sql}
Table~\ref{tab:python-sql-transpilation} reports transpilation accuracy on the 93 questions whose majority-vote answer was produced exclusively by Python across our \spidersnow\ and \spidersqlite\ experiments. The framework achieves 96.77\% accuracy. Manual inspection shows these cases rely on Python features like regular expressions and analytical libraries to build intermediate solutions before conversion to SQL. Our transpilation pipeline uses a two-tier approach, an initial pass with \texttt{gpt-oss-20b}, then a second pass with \texttt{gpt-oss-120b} on remaining failures. The 20b model alone handles $\approx$87\% of cases, covering 40/41 instances in \spidersqlite, while the larger model is needed mainly for Snowflake-specific SQL syntax. The three unsolved examples appear in Appendix~\ref{app:py-to-sql}.

\subsection{Augmenting General-Purpose Coding Agents with \methodabbrev \ Paradigm}
\label{sec:claude_code}

\begin{wraptable}{r}{0.55\textwidth} % 'r' for right, '0pt' for auto-width
% \vspace{em}
\vspace{-1em}
\centering
\resizebox{\linewidth}{!}{
\begin{tabular}{lcc}
\toprule
\textbf{\spidersnow} & \textbf{Pass@1} & \textbf{Average execution time (sec)} \\
\midrule
\textbf{Claude Code} & 57.7 & 216.5 \\ [3pt]
\hdashline
\noalign{\vspace{3pt}}
\quad{with FlexSQL harness} & 66.0 & 288.5 \\
\bottomrule
\end{tabular}
}
\caption{Performance of Claude Code with our custom FlexSQL harness on \spidersnow.}
\label{tab:cc}
\end{wraptable}

As general-purpose coding agents such as Claude Code gain wide adoption, we bring \methodabbrev’s flexible exploration and execution paradigm to Claude Code to evaluate its effectiveness beyond our controlled setting. We package \methodabbrev's tool suite and exploration instructions as Claude skills, encouraging the agent to explore and execute flexibly. As a baseline, we use the same agent equipped with a single query-execution tool. In this experiment, we use Claude Sonnet 4.6 with medium thinking effort, generating one answer per question.

Table \ref{tab:cc} shows our harness performance on \spidersnow. The harness yields an absolute improvement of \textbf{8.3\%} (57.7\% $\rightarrow$ 66.0\% Pass@1) over the baseline. This indicates that flexible exploration and execution are beneficial beyond our original setup, continuing to provide consistent gains when integrated into a widely used general-purpose agent. Though \methodabbrev's performance comes with the cost of increased average latency (216.5 seconds/question $\rightarrow$ 288.5 seconds/question), which mainly due to systematic database exploration and SQL execution, this is a worthwhile trade-off for performance-demanding applications.

\section{Related Work}
\label{sec:related-work}
\paragraph{Text-to-SQL pipelines.}
Text-to-SQL has progressed from early statistical methods \citep{john-mooney-1996, tang-mooney-2000-automated} to language model based systems \citep{text2sqlsurvey, text2sqlsurvey2}, evaluated on increasingly realistic benchmarks \citep{yu2018spider, bird, lei2025spider}.
Most modern approaches follow a fixed pipeline of schema linking, SQL generation, and post-hoc repair \citep{wang-etal-2020-rat, talaei2024chess, deepeyesql, hao2025dsrsql, deng2025reforce, pourreza2025chasesql}, with extensions for planning \citep{sumers2024cognitive, pourreza2023din, mac-sql, li2025alphasql} and candidate diversity \citep{XiYanSQL, lee-etal-2025-mcs, dail_sql}. However, database interactions in these systems are confined to predetermined stages: the schema is retrieved once upfront, and execution feedback is used only for final validation.
\methodabbrev\ instead provides exploration tools throughout reasoning, enabling flexible backtracking when downstream errors reveal incorrect schema assumptions.
 
\paragraph{Agentic reasoning.}
Our work builds on the broader tradition of interleaving reasoning with environment interaction.
ReAct \citep{yao2022react} showed that alternating reasoning traces with actions improves grounding, and subsequent agent frameworks have applied this to code repositories \citep{yang2024sweagent}, operating systems \citep{wu2024copilot}, and interactive coding environments \citep{yang2023intercode}.
Revising or backtracking on earlier decisions based on execution feedback is a well-established strategy in agentic systems \citep{huang2022inner, shinn2023reflexion, madaan2023self, shi2026experiential, song2026expanding}; \methodabbrev\ instantiates this principle for database exploration, where backtracking targets not just code-level errors but also upstream schema and plan assumptions.

\paragraph{Intermediate representations.} Intermediate representations between natural language and SQL have been shown to improve text-to-SQL generation \citep{10.1145/3140587.3062365, 10.1145/3133887, guo-etal-2019-towards, gan-etal-2021-natural-sql} and error correction \citep{chen-etal-2023-text}. Python has emerged as an effective intermediate language for code generation owing to its expressiveness and strong representation in language model training data \citep{transpilation-with-llm, pi-sql}. \methodabbrev\ adopts this idea, allowing the agent to implement solutions in Python and translate them back to SQL.

\section{Conclusion}
\label{sec:conclusion}
We presented \methodabbrev, a text-to-SQL agent that replaces fixed pipeline stages with tools for on-demand database exploration throughout reasoning. This flexibility enables diverse planning, plan-level backtracking, and bilingual generation to each operate over up-to-date schema context, and our ablations confirm that their gains compound when combined.
On \spidertwo, \methodabbrev\ outperforms strong baselines across model scales using open-source models with 20B and 120B parameters, and exhibits effective test-time scaling with increased sampling budgets. Furthermore, integrating our paradigm into general-purpose coding agents like Claude Code yields consistent performance gains, demonstrating the \methodabbrev's architecture generalization across diverse agent platforms.

\section*{Acknowledgments}
This work is supported in part by a Canada CIFAR AI Chair award to XY and a Canada CIFAR AI Chair award to JQC. This research has also been enabled in
part by support provided by the Digital Research Alliance of Canada.

\bibliography{colm2026_conference}
\bibliographystyle{colm2026_conference}

\appendix
\newpage

\section{Building Context}
\label{app:context_building}
Before generating execution plans, \methodabbrev\ performs a preprocessing step to provide the agent with a compact yet informative representation of the database. Following prior work~\citep{deng2025reforce, hao2025dsrsql}, we apply three metadata compression techniques:

\begin{itemize}
    \item \textbf{Null Column Pruning:} We remove columns that contain only NULL values across the entire dataset.
    \item \textbf{Structural Table Grouping:} Tables that share identical schemas but different time-suffixes (e.g., \texttt{GA\_SESSION\_20240101}, \texttt{GA\_SESSION\_20240201}) are grouped together.
    \item \textbf{External Knowledge Extraction:} If external documents are provided, a summarization agent extracts domain knowledge relevant to the query.
\end{itemize}

\section{Generation ratio of Python and SQL}
\label{app:gen_ratio}
Table \ref{tab:py_sql_ratio} shows the breakdown of programming languages (Python and SQL) used within the majority groups of solved questions.   

\begin{table}
\centering
\begin{tabular}{ll ccccc}
\toprule
\multirow{2}{*}{\textbf{Dataset}} & \multirow{2}{*}{\textbf{Model}} & \multirow{2}{*}{\textbf{Only Python}} & \multirow{2}{*}{\textbf{Only SQL}} & \multicolumn{3}{c}{\textbf{Both}} \\
\cmidrule(lr){5-7}
& & & & \textbf{Total} & \textbf{Gen: Py} & \textbf{Gen: SQL} \\
\midrule
\multirow{2}{*}{\textbf{\spidersnow}} & \textbf{120b} & 7.1\%  & 48.8\% & 44.1\% & 37.2\% & 62.8\% \\
 & \textbf{20b} & 5.5\% & 61.0\% & 33.5\% & 31.8\% & 68.2\% \\
\midrule
\multirow{2}{*}{\textbf{\spidersqlite}} & \textbf{120b} & 27.6\% & 25.3\% & 47.1\% & 43.7\% & 56.3\% \\
& \textbf{20b} & 23.3\% & 6.8\% & 69.9\% & 47.7\% & 52.3\% \\
\bottomrule
\end{tabular}%
\caption{Questions solvability breakdown considering the Pass@1 major vote groups on \spidersnow \ and \spidersqlite \ datasets. \emph{Gen: Py} and \emph{Gen: SQL} respectively denotes the portion of Python and SQL programs among the groups.}
\label{tab:py_sql_ratio}
\end{table}

\section{Error Analysis on Python-to-SQL Transpilation}
\label{app:py-to-sql}

We manually inspect three failure cases in Table~\ref{tab:python-sql-transpilation} to understand reasons for unsuccessful Python-to-SQL transpilation. These cases come from two benchmark questions: question sf\_bq194 in \spidersnow\ and question local300 in \spidersqlite.

%% ----------------------------------------------------------------
%%  Case 1: sf_bq194
%% ----------------------------------------------------------------
\paragraph{Case 1: sf\_bq194 (Imperative parsing and language-specific containers).}

This question requires extracting the second most frequently imported library or module across Python (\texttt{.py}), R (\texttt{.r}, \texttt{.Rmd}), and IPython notebook (\texttt{.ipynb}) files. The Python implementation handles each format through separate helper functions, as outlined in the following pseudocode:

\begin{algorithm}
\small
\begin{algorithmic}[1]
\For{each file in dataset}
    \If{extension is \texttt{.py}}
        \State parse import statements line by line
    \ElsIf{extension is \texttt{.ipynb}}
        \State deserialize JSON $\rightarrow$ extract code cells $\rightarrow$ parse imports
    \ElsIf{extension is \texttt{.r} or \texttt{.Rmd}}
        \State match \texttt{library()}/\texttt{require()} and \texttt{::} via regex
    \EndIf
    \State accumulate module names into a frequency counter
\EndFor
\State \Return second most frequent entry from counter
\end{algorithmic}
\end{algorithm}

This logic challenges transpilation for two reasons.
First, each file format requires a fundamentally different parsing strategy: \texttt{.py} files need line-by-line tokenization of \texttt{import} statements, \texttt{.ipynb} files must first be deserialized from JSON before their code cells can be scanned, and \texttt{.r}/\texttt{.Rmd} files rely on regex matching for \texttt{library()} calls and \texttt{::} operators. Encoding all three strategies in SQL would require deeply nested \texttt{CASE} expressions interleaved with dialect-specific string functions, a translation that none of the LLMs attempted correctly.
Second, the program identifies the second most frequent module using \texttt{collections.Counter.most\_common()}, a Python-specific API with no direct SQL counterpart. While counting itself maps naturally to \texttt{GROUP BY} with \texttt{COUNT}, selecting specifically the second-ranked entry requires an additional ranking step (e.g., \texttt{ROW\_NUMBER()} or \texttt{LIMIT}/\texttt{OFFSET}). The LLMs failed to compose this ranking correctly on top of the already complex parsing logic.

%% ----------------------------------------------------------------
%%  Case 2: local300
%% ----------------------------------------------------------------
\paragraph{Case 2: local300 (Stateful recurrence and multi-stage aggregation).}

This question asks to compute daily balances over each customer's full transaction period, clamping negative balances to zero, and then report, for each month, the sum of every customer's highest daily balance in that month. The central difficulty is that the daily balance is defined by a recurrence:
\begin{equation}
\label{eq:balance-recurrence}
\mathsf{balance}[t] = \max\!\bigl(\mathsf{balance}[t{-}1] + \mathsf{amount}[t],\; 0\bigr)
\end{equation}

The $\max(\cdot, 0)$ clamp makes this fundamentally different from a cumulative sum. To illustrate, consider a customer with daily transaction amounts $[+5, -8, +3]$. A cumulative sum would yield running balances of $[5, -3, 0]$, whereas the recurrence in Eq.~\eqref{eq:balance-recurrence} produces $[5, 0, 3]$: the balance resets to zero on day~2, so day~3 starts from zero rather than from~$-3$. Because each day's output depends on the \emph{clamped} result of the previous day, this cannot be expressed with a \texttt{SUM() OVER (ORDER BY date)} window function; instead, it requires a \textbf{recursive CTE} that materializes one row per customer per day, carrying forward the clamped balance.

In all three transpilation attempts, the LLMs used a window-function-based cumulative sum, producing the $[5, -3, 0]$ trajectory rather than the correct $[5, 0, 3]$.
The task is further complicated by multi-stage aggregation that follows the recurrence: daily balances must be grouped by customer and month to extract each customer's monthly maximum, and those maxima must then be summed across all customers. Each stage introduces a different grouping key and aggregation target, and the LLMs struggled to keep these consistent throughout the translation.

\section{Examples of Python's advantages on \spidersnow \ analytical questions}

While Snowflake SQL can express complex analytics through specialized extensions, LLMs can generate more reliable solutions in Python for certain classes of analytical queries. This advantage stems from two factors. First, Python's procedural paradigm aligns naturally with LLMs' autoregressive generation: complex reasoning decomposes into sequential steps, whereas SQL demands that the same logic be expressed declaratively in a single, often heavily nested statement. Second, Python's rich data processing libraries (e.g., \texttt{pandas}, \texttt{numpy}) is prevalent in LLM training data \citep{transpilation-with-llm, twist2025study}, yielding strong out-of-the-box proficiency. The following examples illustrates where these advantages are particularly clear.

\textbf{\texttt{sf\_local279}} requires simulating a monthly inventory system throughout 2019, where each month's ending stock depends on the previous month's result and a conditional restocking rule: if inventory falls below a product-specific minimum, a fixed purchase quantity is added. The task then asks for the month per product where ending inventory is closest to the minimum threshold. In Python, this is a straightforward loop with an accumulator variable and an \texttt{if}/\texttt{else} branch, naturally expressing the sequential state dependency. In SQL, the same logic demands a recursive CTE that propagates inventory forward month by month, embedding the conditional restock decision within a \texttt{CASE WHEN} expression and carrying mutable state as a column across recursive iterations. Crucially, the recursion cannot be replaced by a window function, because each month's restock decision alters all subsequent inventory levels, making the running state path-dependent rather than purely cumulative.

\textbf{\texttt{sf\_local002}} asks the system to forecast toy sales for four dates in December 2018 by fitting a simple linear regression to historical daily sales and then computing five day symmetric moving averages over the predicted values. In Python, pandas supports date range generation and aggregation, while \texttt{numpy.polyfit} fits the regression in a single call, and the rolling window can be applied directly using \texttt{pandas.rolling}. Snowflake SQL can express this using built in \texttt{REGR\_SLOPE} and \texttt{REGR\_INTERCEPT} aggregate functions together with a recursive CTE for calendar generation. However, functions like \texttt{REGR\_SLOPE} are platform specific extensions, and their availability and syntax vary across database systems. Allowing a language model to flexibly generate in Python gives it a broader set of tools to draw from, which could lead to better outcomes on tasks involving non standard analytical operations.

\section{Statistics of tool calls}
Table \ref{tab:tool_call_counts} details the number of tool calls during our Pass@8 experiments. Generally, the language models prefer the two code execution tools due to their inherent flexibility. Among the tools,\texttt{GetSchema} is exclusive to data warehouse systems like Snowflake, where tables are organized into schemas (Section \ref{sec:tools}), and only 144 questions in \spidersnow\ require accessing multiple schemas, accounting for its low count relative to other tools.

\begin{table}[t]
\centering
\begin{tabular}{lrrrr}
\toprule
\textbf{Tool} & \textbf{20b-snow} & \textbf{120b-snow} & \textbf{20b-lite} & \textbf{120b-lite} \\
\midrule
\texttt{GetSchema}      & 280    & 251    & --     & --     \\
\texttt{GetColumns}     & 18,538 & 27,355 & 7,295  & 8,649  \\
\texttt{GetColValues}   & 3,083  & 6,553  & 565    & 1,131  \\
\texttt{FindRows}       & 4,880  & 9,400  & 292    & 677    \\
\texttt{SQLExecutor}    & 37,261 & 44,297 & 4,838  & 5,138  \\
\texttt{PythonExecutor} & 21,996 & 27,051 & 24,303 & 9,194  \\
\midrule
\textbf{Total} & \textbf{86,038} & \textbf{114,907} & \textbf{37,293} & \textbf{24,789} \\
\bottomrule
\end{tabular}
\caption{Number of tool calls for \texttt{gpt-oss-120b} and \texttt{gpt-oss-20b} in our experiments with Pass@8 on \spidersnow \ and \spidersqlite.}
\label{tab:tool_call_counts}
\end{table}

% \begin{table}[t]
% \centering
% \resizebox{\linewidth}{!}{%
% \begin{tabular}{lccccccc}
% \toprule
%  & \multicolumn{4}{c}{Pass@16} & \multirow{2}{*}{Pass@1} & \multirow{2}{*}{Pass@1 (MV)} & \multirow{2}{*}{Pass@1 (MV) Strict} \\
% \cmidrule(lr){2-5}
%  & Simple & Moderate & Challenging & Total & & & \\
% \midrule
% \textbf{Ours} & \textbf{97.06} & \textbf{87.88} & \textbf{75.75} & \textbf{87.00} & \textbf{68.00} & 75.00 & 40.00 \\
% DSR-SQL & 91.18 & 72.73 & 72.73 & 79.00 & 54.00 & 78.00 & 43.00 \\
% Alpha-SQL\textsuperscript{*} & - & - & - & - & - & 75.00 & 70.00 \\
% \bottomrule
% \end{tabular}
% }
% \caption{Our framework vs \dsrsql \ performance on BIRD with \texttt{gpt-oss-20b}. \textsuperscript{*}From \cite{li2025alphasql}'s generated queries using Qwen2.5-Coder-32B-Instruct. \emph{Pass@1 (MV) Strict} denotes result using BIRD's exact table matching evaluation script.}
% \label{tab:bird_subset_result}
% \end{table}

\section{Heuristic rules for transpilation}
\label{app:heuristic-transpilation-rules}

Our heuristic rules for transpiling Python to SQL are summarized as follows:

\begin{enumerate}[leftmargin=*, itemsep=4pt]

    \item \textbf{Runtime Type Preservation.}
    If the Python program's behavior depends on runtime types (e.g., \texttt{isinstance(x, str)}, \texttt{dict}, \texttt{list}, \texttt{tuple}), the SQL translation must preserve that behavior. For instance, do not use string operations unless the runtime value is a string.

    \item \textbf{VARIANT Column Unnesting.}
    The SQL translation can handle (semi-)structured data types (e.g., Snowflake's \texttt{VARIANT} and nested objects/arrays). When a Python code snippet defines a helper function that:
    \begin{itemize}[nosep, leftmargin=3em]
        \item[$-$] takes a single column value as input,
        \item[$-$] handles null\,/\,missing values by returning an empty list,
        \item[$-$] parses a JSON string or \texttt{VARIANT}-like value using \texttt{json.loads},
        \item[$-$] converts a JSON array or dictionary into a Python list, and
        \item[$-$] is applied row-wise via \texttt{Series.apply}, \texttt{map}, or equivalent,
    \end{itemize}
    this snippet is unnesting a Snowflake \texttt{VARIANT} column and should be translated into:\\[2pt]
    \centerline{\texttt{(LATERAL) FLATTEN(input => COLUMN \textbar\ PARSE\_JSON(COLUMN))}}

    \item \textbf{Identifier vs.\ String Distinction.}
    The SQL translation must clearly distinguish quoted identifiers from strings and regular expressions.
    
    \item \textbf{Regex Semantics Alignment.}
    Python and Snowflake use different regex semantics. When translating \texttt{str.contains(REGEX, case=\textit{bool})} to SQL, use:
    \[
        \texttt{REGEXP\_LIKE(str,\ '.*'\ ||\ REGEX\ ||\ '.*',\ \textit{flags})}
    \]
    You \emph{must} prepend and append \texttt{'.*'} to \texttt{REGEX} to ensure semantic equivalence. If \texttt{case=False}, use the flag \texttt{'is'} (not \texttt{'i'} alone) in \texttt{REGEXP\_LIKE} to ensure consistency with Python's regex behavior.
    \item \textbf{VARIANT Data Access Paths.}
    When writing SQL queries that access \texttt{VARIANT} columns in Snowflake (i.e., columns where each value is a JSON object), you must specify the full access path, including nested keys and array indices:
    \begin{itemize}[nosep, leftmargin=3em]
        \item[$-$] use single quotes for case-sensitive key access, e.g., \texttt{"col":'KeyName'},
        \item[$-$] chain keys for nested object access, e.g., \texttt{"col":'OuterKey':'InnerKey'},
        \item[$-$] use zero-based indexing for array element access, e.g., \texttt{"col"[0]},
        \item[$-$] use \texttt{GET\_PATH("col", "dynamic\_key\_string")} for dynamic key access.
    \end{itemize}
    Always include the \emph{full} path to the target value (e.g., \texttt{"col":'key':'sub'[0]}) rather than just the base column name.

    \item \textbf{JSON String Columns.}
    For columns whose values are JSON-formatted strings, first apply \texttt{PARSE\_JSON()} to convert the string into a JSON object, then use the access rules in Rule~5.

\end{enumerate}

\section{Inference Configurations}
\label{sec:gen_config}
In all of our experiments, we set up local vLLM instances \citep{kwon2023efficient} using at most 4 Nvidia L40s GPUs to serve \texttt{gpt-oss} models \citep{gptoss} with \texttt{"reasoning\_effort" = "high"} and \texttt{"temperature" = 1.0}.

\section{Additional discussions}

\paragraph{Evaluation scope.}
Our main evaluation focuses on \spidertwo\ due to its structural complexity and and strong grounding in enterprise-level databases. While foundational benchmarks such as Spider \citep{yu2018spider} remain valuable for evaluating standard text-to-SQL generation, their smaller-scale environments abstract away many real-world challenges, making them less indicative of how agentic methods perform in practice. By contrast, \spidertwo\ introduces massive schemas and dialect-specific features that more effectively stress-test advanced reasoning and planning.

\paragraph{Comparison with closed-source pipelines.} To ensure reproducibility and facilitate controlled comparisons under a shared model setting, we restrict our baselines to open-source approaches with publicly available code and literature. We do not benchmark against several top entries on the \spidertwo\ leaderboard that rely on proprietary models or undisclosed system designs. The limited transparency regarding their underlying architectures, prompting strategies, and compute budgets makes it difficult to establish fair baseline settings for comparison.

\end{document}